\documentclass[graybox,envcountsect]{svmult}

\usepackage{mathptmx}       % selects Times Roman as basic font
\DeclareMathAlphabet{\mathcal}{OMS}{cmsy}{m}{n}
\usepackage{helvet}         % selects Helvetica as sans-serif font
\usepackage{courier}        % selects Courier as typewriter font
\usepackage{type1cm}        % activate if the above 3 fonts are
\usepackage{makeidx}         % allows index generation
\usepackage{graphicx}        % standard LaTeX graphics tool
\usepackage{multicol}        % used for the two-column index
\usepackage[bottom]{footmisc}% places footnotes at page bottom

\usepackage{amssymb}
\usepackage{url}
\usepackage{verbatim}

\usepackage{amsmath}

\usepackage{graphicx}

\usepackage{algorithmic}
\usepackage[ruled,vlined]{algorithm2e}
\usepackage{amsfonts}
\usepackage{natbib}
\usepackage[colorlinks=true,linkcolor=blue, citecolor=blue]{hyperref}
\usepackage{xcolor}
\usepackage{array}
\usepackage{multirow}
\usepackage{breakcites}
\usepackage[T1]{fontenc}

\newcommand{\R}{\mathbb{R}}
\newcommand{\EX}{\mathbb{E}}
\newcommand{\pip}{\pi_{\mathrm{\widetilde{P}}}}
\newcommand{\pin}{\pi_{\mathrm{\widetilde{N}}}}
\newcommand{\EJoint}{\mathop{\mathbb{E}}\limits_{(\x,y) \sim \mathcal{D}}}
\newcommand{\Epos}{\mathbb{E}_\mathrm{P}}
\newcommand{\Eposc}{\mathbb{E}_\mathrm{\widetilde{P}}}
\newcommand{\Eneg}{\mathbb{E}_\mathrm{N}}
\newcommand{\Enegc}{\mathbb{E}_\mathrm{\widetilde{N}}}

\newcommand{\zerooneloss}{\ell_{0\text{-}1}}

\newcommand{\x}{\boldsymbol{x}}
\newcommand{\X}{\mathcal{X}}

\newcommand{\sell}{\ell_\mathrm{sym}}

\newcolumntype{L}{>{$}l<{$}}
\newcolumntype{C}{>{$}c<{$}}

\makeindex             % used for the subject index
\begin{document}

\title*{\bf A Symmetric Loss Perspective of Reliable Machine Learning}

\author{Nontawat Charoenphakdee, Jongyeong Lee, Masashi Sugiyama}

\institute{Nontawat Charoenphakdee \at
              The University of Tokyo, RIKEN AIP \\
            \email{nontawat@ms.k.u-tokyo.ac.jp}           %  \\
          \and
          Jongyeong Lee \at
              The University of Tokyo, RIKEN AIP \\
            \email{lee@ms.k.u-tokyo.ac.jp} 
            \and
          Masashi Sugiyama \at
            RIKEN AIP, The University of Tokyo \\
            \email{sugi@k.u-tokyo.ac.jp}
}


\maketitle
\abstract{
When minimizing the empirical risk in binary classification, it is a common practice to replace the zero-one loss with a surrogate loss to make the learning objective feasible to optimize. 
Examples of well-known surrogate losses for binary classification include the logistic loss, hinge loss, and sigmoid loss. 
It is known that the choice of a surrogate loss can highly influence the performance of the trained classifier and therefore it should be carefully chosen. 
Recently, surrogate losses that satisfy a certain symmetric condition (aka., symmetric losses) have demonstrated their usefulness in learning from corrupted labels.
In this article, we provide an overview of symmetric losses and their applications.
First, we review how a symmetric loss can yield robust classification from corrupted labels in balanced error rate (BER) minimization and area under the receiver operating characteristic curve (AUC) maximization.
Then, we demonstrate how the robust AUC maximization method can benefit natural language processing in the problem where we want to learn only from relevant keywords and unlabeled documents.
Finally, we conclude this article by discussing future directions, including potential applications of symmetric losses for reliable machine learning and the design of non-symmetric losses that can benefit from the symmetric condition.  
}


\section{Introduction}
\label{intro}
Modern machine learning methods such as deep learning typically require a large amount of data to achieve desirable performance~\citep{schmidhuber2015deep,lecun2015deep,goodfellow2016deep}.
However, it is often the case that the labeling process is costly and time-consuming.
To mitigate this problem, one may consider collecting training labels through crowdsourcing~\citep{dawid1979maximum,kittur2008crowdsourcing}, which is a popular approach and has become more convenient in the recent years~\citep{deng2009imagenet,crowston2012amazon,sun2014chimera,vaughan2017making,pandey2020crowdsourcing,vermicelli2020can, washington2020precision}.
For example, crowdsourcing has been used for tackling the COVID-19 pandemic to accelerate research and drug discovery~\citep{vermicelli2020can,chodera2020crowdsourcing}.
However, a big challenge of crowdsourcing is that the collected labels can be unreliable because of non-expert annotators fail to provide correct information~\citep{lease2011quality,zhang2014spectral,gao2016exact,imamura2018analysis}. 
Not only the non-expert error, but even expert annotators can also make mistakes.
As a result, it is unrealistic to always expect that the collected training data are always reliable.  

It is well-known that training from data with noisy labels can give an inaccurate classifier~\citep{noise1, noise2, noise3,frenay2013classification,natarajan2013learning}.
Interestingly, it has been shown that the trained classifier may only perform slightly better than random guessing even under a simple noise assumption~\citep{long2010random}.
Since learning from noisy labels is challenging and highly relevant in the real-world, this problem has been studied extensively in both theoretical and practical aspects~\citep{van2017theory,jiang2018mentornet,algan2019image,liu2020peer,wei2020optimizing,karimi2020deep,han2018co,han2020survey}.

Recently, a loss function that satisfies a certain symmetric condition has demonstrated its usefulness in learning from noisy labels.
A pioneer work in this direction is the work by~\citet{manwani2013noise}, where they showed that using symmetric losses can be robust under random classification noise~(see also~\citet{ghosh2015making} and~\citet{van2015learning}).
However, the assumption of random classification noise can be restrictive since it assumes that each training label can be flipped independently with the fixed probability regardless of its original label.
As a result, it is important to investigate a more realistic noise model that reflects a real-world situation more accurately.  
In this article, we review the robustness result of symmetric losses in the setting of mutually contaminated noise~\citep{scott2013classification,menon2015}.
This noise model has been proven to be quite general since
it encompasses well-known noise assumptions such as the random classification noise and class-conditional noise~\citep{menon2015,lu2018minimal}.
Furthermore, many instances of weakly supervised learning problems can also be formulated into the setting of mutually contaminated noise~\citep{kiryo2017,baosu,lu2018minimal,shimada2019classification, sugiyama2022machine}.
In this article, we will discuss how using a symmetric loss can be advantageous in BER and AUC optimization under mutually contaminated noise.
With a symmetric loss, one does not need the knowledge of the noise rate to learn effectively with a theoretical guarantee~\citep{charoenphakdee2019symmetric}.

This article also demonstrates how to use a symmetric loss in a real-world problem in the context of natural language processing.
We discuss an application of symmetric losses for learning a reliable classifier from only relevant keywords and unlabeled documents~\citep{jin2017combining,charoenphakdee2019learning}.
In this problem, we first collect unlabeled documents. 
Then, we collect relevant keywords that are useful for determining the target class of interest.
Unlike collecting labels for every training document, collecting keywords can be much cheaper and the number of keywords does not necessarily scale with the number of unlabeled training documents~\citep{chang2008importance,song2014dataless,chen2015dataless,li2018pseudo,jin2017combining,jin2020learning}.
We will discuss how this problem can be formulated into the framework of learning under mutually contaminated noise and how using a symmetric loss can be highly useful for solving this problem~\citep{charoenphakdee2019learning}. 
 
% The rest of this article is organized as follows.
% Section~\ref{sec:prelim} presents the background of binary classification from corrupted labels and evaluation metrics that are considered in this paper.
% Section~\ref{sec:symadvantage} illustrates the advantage of using symmetric losses for BER and AUC optimization from corrupted labels.
% Section~\ref{sec:application} demonstrates an application of symmetric losses for a natural language processing task, which is the task of learning only from relevant keywords and unlabeled documents.
% Finally, we conclude this article by discussing other potential applications of symmetric losses and future direction in Section~\ref{sec:discussion}.

\section{Preliminaries}
\label{sec:prelim}
In this section, we review the standard formulation of binary classification based on empirical risk minimization~\citep{vapnik1998statistical} and well-known evaluation metrics.
Then, we review the definition of symmetric losses and the problem of learning from corrupted labels.  

\subsection{Binary classification}
\label{sec:binclass}
Here, we review the problem of binary classification, where the goal is to learn an accurate classifier from labeled training data.

Let $\x \in \X$ denote a pattern in an input space $\X$.
For example, an input space $\X$ could be a space of $d$-dimensional real-valued vectors in $\R^d$.
Also, let $y \in \{-1, +1\}$ denote a class label and $g\colon\X \to \R$ denote a prediction function that we want to learn from data. 
In binary classification, we use the function sign$(g(\x))$ to determine the predicted label of a prediction function, where sign$(g(\x))=1$ if $g(\x)>0$, $-1$ if $g(\x)<0$, and $0$ otherwise\footnote{$\mathrm{sign}(g(\x))=0$ indicates that $g$ suggests to randomly guess a label. In practice, one may force a prediction function $g$ to not output zero when training $g$. For simplicity of explanation, we assume that $g$ does not output zero.} .
In ordinary binary classification, we are given $n$ labeled training examples, $\{\x_i, y_i\}_{i=1}^{n}$,
which are assumed to be drawn independently from a joint distribution $\mathcal{D}$ with density $p(\x, y)$. 
Next, to evaluate the prediction function, we define the zero-one loss as follows:
\begin{align}
    \zerooneloss(z)=-\frac{1}{2} \, \mathrm{sign}(z) + \frac{1}{2}.
\end{align}
Given an input-output pair $(\x,y)$, if the signs of both $y$ and $g(\x)$ are identical, we have zero penalty, i.e., $\zerooneloss(yg(\x))=0$. 
On the other hand, we have $\zerooneloss(yg(\x))=1$ if the signs of $y$ and $g(\x)$ are different, which indicates incorrect prediction.

The goal of binary classification is to find a prediction function $g$ that minimizes the following misclassification risk, i.e., the risk corresponding to the classification error rate (CER):
\begin{align}\label{pnrisk}
R^{\zerooneloss}_{\mathrm{CER}}(g) = \EJoint \left[ \zerooneloss(y g(\x) )\right].
\end{align}

As suggested by Eq.~\eqref{pnrisk}, our goal is to find the prediction function that performs well w.r.t. the whole distribution on average, that is, the prediction function $g$ should be able to also classify unseen examples from the same data distribution accurately, not only perform well on the observed training examples.

In practice, the misclassification risk $R^{\zerooneloss}_{\mathrm{CER}}$ cannot be directly minimized because we only observe finite training examples, not the whole probability distribution.
By using training examples, the empirical risk minimization framework suggests to find~$g$ by minimizing the following empirical risk~\citep{vapnik1998statistical}:
\begin{align}\label{pnriskemp}
\widehat{R}^{\zerooneloss}_{\mathrm{CER}}(g) = \frac{1}{n} \sum_{i=1}^{n} \zerooneloss(y_ig(\x_i))  \text{.}
\end{align}

Although the risk estimator in Eq.~\eqref{pnriskemp} is an unbiased and consistent estimator of the misclassification risk~\citep{vapnik1998statistical}, it is not straightforward to directly minimize it.
Indeed, with the zero-one loss in the empirical risk, the minimization problem is known to be computationally infeasible. It is an NP-hard problem even if the function class to search $g$ is a class of linear hyperplanes~\citep{zeroonenphard2, zeroonenphard1}.
Moreover, the gradient of the zero-one loss is zero almost everywhere and therefore hinders the use of a gradient-based optimization method.
To mitigate this problem, it is a common practice to replace the zero-one loss with a different loss function $\ell$ that is easier to minimize, which is called a \emph{surrogate loss}~\citep{zhang2004statistical,bartlett2006}.
As a result, we minimize the following empirical surrogate risk:
\begin{align}
\widehat{R}_{\mathrm{CER}}^\ell(g) = \frac{1}{n} \sum_{i=1}^{n}  \ell(y_ig(\x_i)),
\end{align}
where regularization techniques can also be employed to avoid overfitting.

\begin{table*}[t]
\centering
\caption{Classification-calibrated losses and their properties including the convexity and whether they satisfy the symmetric condition $\ell(z)+\ell(-z)=K$, where $K$ is a constant.} \label{table:optimal-loss}
\begin{tabular}{|C|C|C | C |}
\hline
\text{Loss} & \ell(z)  & \text{Convex} & \text{Symmetric} \\ \hline
\text{Zero-one} & -\frac{1}{2} \, \mathrm{sign}(z) + \frac{1}{2}&  \times &\checkmark\\ 
\text{Squared} & (1-z)^{2} &  \checkmark &\times\\ 
\text{Hinge} & \max(0, 1-z) &  \checkmark &\times \\ 
% \text{Perceptron} & \max(0, -z) &  \checkmark &\times \\ 
\text{Squared hinge} & \max(0, 1-z)^{2} &  \checkmark &\times \\ 
\text{Exponential} & \exp(-z)&  \checkmark &\times\\ 
\text{Logistic} & \mathrm{log}(1+\exp(-z)) & \checkmark &\times \\ 
\text{Savage}  & \left[(1+\exp(2z))^{2}\right]^{-1} &\times &  \times \\ 
\text{Tangent} & (2\mathrm{arctan}(z)-1)^{2}  &\times &  \times\\
\text{Ramp}& \mathrm{max}(0, \mathrm{min}(1, (1-z)/2)) & \times &  \checkmark \\ 
 \text{Sigmoid} & \left[1+\exp(z)\right]^{-1} &\times &  \checkmark \\ 
  \text{Unhinged} & 1-z &\checkmark &  \checkmark \\ 
     \hline
\end{tabular}
\end{table*}

The choice of a surrogate loss $\ell$ is highly crucial for training a good classifier and should be carefully designed. 
This is because the ultimate goal is still to minimize the misclassification risk $R_{\mathrm{CER}}^{\zerooneloss}(g)$, not the surrogate risk $R_{\mathrm{CER}}^{\ell}(g)$.
To ensure that minimizing the surrogate risk $R_{\mathrm{CER}}^\ell(g)$ yields a meaningful solution for the misclassification risk $R_{\mathrm{CER}}^{\zerooneloss}(g)$, a surrogate loss should satisfy a \emph{classification-calibration} condition, which is known to be a minimum requirement for binary classification~(see~\citet{bartlett2006} for more details).
Many well-known surrogate losses in binary classification satisfy this property.
Table~\ref{table:optimal-loss} provides examples of classification-calibrated losses and their properties~\citep{bartlett2006,masnadi2009,masnadi2010design,van2015learning}.

\subsection{Beyond classification error rate}
Although CER has been used extensively, one should be aware that using this evaluation metric may not be informative when the test labels are highly imbalanced~\citep{menon2013statistical}.
For example, consider a trivial prediction function $g_\mathrm{pos}$ such that $g_\mathrm{pos}(\x) >0$ for any $\x$, that is, $g_\mathrm{pos}$ only predicts a positive label. 
If $99\%$ of the test labels are positive, CER of $g_\mathrm{pos}$ is $0.01$, which may indicate very good performance. 
However, $g_\mathrm{pos}$ does not give any meaningful information since it always predicts a positive label regardless of an input $\x$. 
Thus, low CER may mislead someone into thinking that $g_\mathrm{pos}$ is a good classifier. 
Here, we review evaluation metrics that can be used as an alternative to CER to prevent such a problem.
\subsubsection{Balanced error rate (BER)}
Let $\Epos[\cdot] $ and $\Eneg[\cdot]$ be the expectations of $\x$ over $p(\x|y=+1)$ and $p(\x|y=-1)$, respectively. 
Then, the BER risk is defined as follows:
\begin{align*}
R^{\zerooneloss}_{\mathrm{BER}}(g) &= \frac{1}{2} \bigg[ \Epos\left[{\zerooneloss}(g(\x))\right] + \Eneg\left[{\zerooneloss}(-g(\x))\right] \bigg] \text{.}
\end{align*}
It is insightful to note that CER can also be expressed as
\begin{align*}
    R^{\zerooneloss}_{\mathrm{CER}}(g) &=  p(y=+1)\Epos\left[{\zerooneloss}(g(\x))\right] + (1-p(y=+1))\Eneg\left[{\zerooneloss}(-g(\x))\right]  \text{,}
\end{align*}
where $p(y=+1)$ is the class prior given by $p(y=+1)= \int p(\x, y=+1) \, \mathrm{d}\x$. We can see that the BER minimization problem is equivalent to the CER minimization problem if the class prior is balanced, i.e., $p(y=+1)=\frac{1}{2}$.
Furthermore, unlike $R^{\zerooneloss}_{\mathrm{CER}}$, any trivial prediction function that predicts only one label cannot have $R^{\zerooneloss}_{\mathrm{BER}}$ lower than $\frac{1}{2}$ regardless of the class prior.
As a result, the prediction function $g$ has an incentive to predict both classes to obtain a low balanced error risk $R^{\zerooneloss}_{\mathrm{BER}}$.
Therefore, BER is known to be useful to evaluate the prediction function $g$ under class imbalance~\citep{ber2,ber1,brodersen2010balanced}. 
In addition, it is also worth noting that BER can be interpreted as an arithmetic mean of the false positive and false negative rates~\citep{menon2013statistical}.

Similarly to the CER minimization problem, we can minimize the empirical surrogate risk using training data and a classification-calibrated loss as follows: 
\begin{align}
    \widehat{R}^{\ell}_{\mathrm{BER}}(g) = \frac{1}{2} \left[\frac{1}{n_\mathrm{P}}\sum_{i: y_i=+1}\ell(g(\x_i)) + \frac{1}{n_\mathrm{N}}\sum_{j: y_j=-1}\ell(-g(\x_j)) \right],
\end{align}
where $n_\mathrm{P}$ and $n_\mathrm{N}$ are the numbers of positive and negative examples, respectively.

\subsubsection{Area under the receiver operating characteristic curve (AUC)}
In binary classification, a receiver operating characteristic (ROC) curve plots the true positive rate against the false positive rate at various decision thresholds.
The area under the ROC curve is called the AUC score, which can be used to evaluate the performance of a prediction function over all possible decision thresholds on average.
The AUC score can also be interpreted as the probability that the prediction function outputs a higher score for a random positive example than a random negative example~\citep{fawcett2006introduction}.

Let us consider the following AUC risk~\citep{narasimhan2013relationship}:
\begin{align} \label{aucrisk}
R^{\zerooneloss}_{\mathrm{AUC}}(g) = \Epos [ \Eneg[{\zerooneloss}( g(\x_\mathrm{P})-g(\x_\mathrm{N}))]] \text{,}
\end{align}
which is the complement of AUC score since the expected AUC score is $1-~R^{\zerooneloss}_{\mathrm{AUC}}(g)$. 
Therefore, maximizing the AUC score is equivalent to minimizing the AUC risk. Intuitively, the high AUC score indicates that $g$ outputs a higher value to positive examples than negative examples on average.
Unlike CER and BER where the function $\mathrm{sign}(g(\x))$ is crucial for the evaluation, in AUC, the sign function is evaluated on the difference between the outputs of $g$ on positive and negative data.   
As a result, an evaluation based on AUC is highly related to the bipartite ranking problem~\citep{narasimhan2013relationship,menon2016bipartite}, where the goal is to find a function $g$ that can rank positive examples over negative examples.
It is also worth noting that AUC is highly related to the Wilcoxon-Mann-Whitney
statistic~\citep{mann1947test,hanley1982meaning}.
Similarly to BER, AUC is also known to be a useful evaluation metric under class imbalance~\citep{ber2, ber1}.

Given training data, the empirical surrogate AUC risk can be defined as follows:
\begin{align}
    \widehat{R}^{\ell}_{\mathrm{AUC}}(g)  =  \frac{1}{n_\mathrm{P}\times n_\mathrm{N}}\sum_{i:y_i=+1} \sum_{j:y_j=-1}\ell(g(\x_i) - g(\x_j)).
\end{align}
However, unlike the CER and BER minimization problems, a loss requirement for AUC optimization should be \emph{AUC-consistent} to guarantee that the optimal solution of a surrogate AUC risk  is also optimal for the AUC risk~\citep{gao2015consistency,menon2016bipartite}. 
Note that this condition is not equivalent to classification-calibration. 
For example, the hinge loss is known to be classification-calibrated but not AUC-consistent~\citep{gao2015consistency,uematsu2017theoretically}.

\subsection{Symmetric losses}
\begin{figure}[t]
  \centering
\includegraphics[width=\columnwidth]{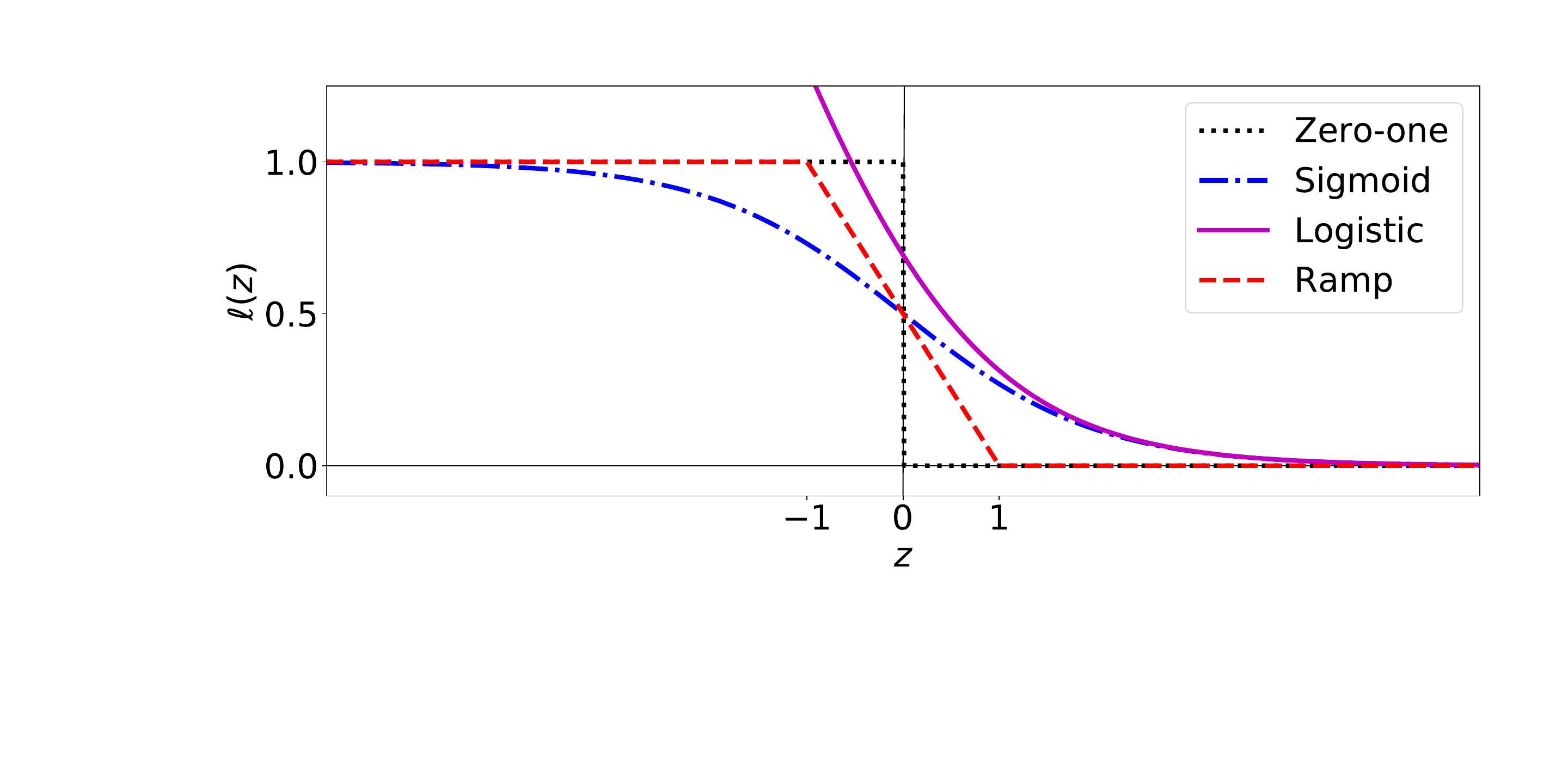}
\vspace{-0.5in}
  \caption{Examples of loss functions. 
  All losses in this figure except the logistic loss are symmetric.}
  \label{fig:all-loss-ex}
\end{figure}

The term symmetric loss in this article refers to a loss function $\ell_\mathrm{sym}: \X \to \R$ such that $\ell_\mathrm{sym}(z)+\ell_\mathrm{sym}(-z) = K$, where $K$ is a constant\footnote{There are other definitions of symmetric loss. For example, see ~\citet{reid2010} and \citet{natarajan2013learning} for other definitions.}. 
It is known that if a loss function is symmetric and non-negative, it must be non-convex~\citep{du2014}.
Figure~\ref{fig:all-loss-ex} illustrates three symmetric losses, which are the zero-one loss, sigmoid loss, and ramp loss.
It is worth noting that both the sigmoid loss and the ramp loss are classification-calibrated and AUC-consistent~\citep{charoenphakdee2019symmetric}. 

\subsection{Learning from corrupted labels}~\label{sec:learncor}
The standard formulation of binary classification in Section~\ref{sec:binclass} does not take into account that training labels can be corrupted.
To extend the standard formulation to support such a situation, learning from corrupted labels under mutually contaminated noise assumes that the training data are given as follows~\citep{menon2015,lu2018minimal,charoenphakdee2019symmetric}:
\begin{align*}
 X_\mathrm{\widetilde{P}}&:= \{\x^\mathrm{\widetilde{P}}_i\}_{i=1}^{n_\mathrm{\widetilde{P}}} \stackrel{\mathrm{i.i.d.}}{\sim} p_{\pip}(\x) \text{,}\\
 X_{\mathrm{\widetilde{N}}}&:=  \{\x^\mathrm{\widetilde{N}}_j\}_{j=1}^{n_\mathrm{\widetilde{N}}}\stackrel{\mathrm{i.i.d.}}{\sim} p_{\pin}(\x) \text{,}
\end{align*}
where, for $0<\pin<\pip<1$,
\begin{align*}
p_{\pip}(\x)&= \pip p(\x|y=+1)+(1-\pip) p(\x|y=-1) \text{,}\\
 p_{\pin}(\x)&=  \pin p(\x|y=+1)+(1-\pin) p(\x|y=-1) \text{.}
\end{align*}
%``$\widetilde{P}$'' and ``$\widetilde{N}$" denote ``corrupted positive" and ``corrupted negative", respectively.

Concretely, the formulation assumes that corrupted positive examples $X_\mathrm{\widetilde{P}}$ are drawn from the distribution where its density is a mixture of the class-conditional positive density $p(\x|y=+1)$ and class-conditional negative density $p(\x|y=-1)$, where $\pip$ controls the mixture proportion between two densities.
Corrupted negative examples $X_\mathrm{\widetilde{N}}$ are also assumed to be drawn similarly but with a different mixture proportion $\pin$.

We can interpret the assumption of the data  generating process as follows. 
The given training data is clean if $\pip=1$ and $\pin=0$, i.e., the data of each class is drawn from the class-conditional distribution w.r.t. one class. 
On the other hand, we can see that the training data can be highly noisy if $\pip-\pin$ is small, i.e., the corrupted positive data and corrupted negative data have a similar distribution and therefore difficult to distinguish.
Note that in this framework, it is reasonable to assume that $\pip > \pin$ because the corrupted positive distribution should still have more information about positive data than the corrupted negative distribution.

In learning from corrupted labels, it has been shown that CER minimization based on empirical risk minimization is possible if the knowledge of $\pip$, $\pin$, and the class prior of the test distribution is given~\citep{lu2018minimal,lu2020mitigating}. 
On the other hand, the problem of estimating $\pip$ and $\pin$ from corrupted labeled data is known to be an unidentifiable problem unless a restrictive condition is applied~\citep{blanchard2010semi, menon2015, scott2015rate}.
Furthermore, the class prior of the test distribution has no relationship with $\pip$ and $\pin$ and it has to be specified if the goal is to minimize the misclassification risk~\citep{lu2018minimal}.
For these reasons, CER minimization can be infeasible if one has access only to corrupted labeled examples without any additional information.
As a result, it is important to explore other evaluation metrics that could be useful in learning from corrupted labels that can be optimized \emph{without requiring the knowledge of $\pip$ and $\pin$}.
In Section~\ref{sec:symadvantage}, we will show that BER and AUC can be effectively optimized in learning from corrupted labels without having to estimate $\pip$ and~$\pin$.
In Table~\ref{table:references}, we show related work on learning from clean and corrupted labels with different evaluation metrics that could be of interest to readers.

\begin{table*}[t]
\centering
\caption{References of related work on evaluation metrics and problem settings mentioned in this article. 
Section~\ref{sec:symadvantage} of this article focuses on the problem of BER and AUC optimization from corrupted labels under mutually contaminated noise, which is based on~\citet{charoenphakdee2019symmetric}.}
\label{table:references}
\begin{tabular}{|c|p{0.35\linewidth} | p{0.35\linewidth} |}
\hline
    & \multicolumn{1}{c|}{\centering Clean labels} &  \multicolumn{1}{c|}{\centering Corrupted labels}              \\ \hline
CER &   \citet{zhang2004statistical,bartlett2006,ben2012minimizing}         &          \citet{lu2018minimal,lu2020mitigating}                     \\ \hline
BER &    \citet{brodersen2010balanced,menon2013statistical}          & \multicolumn{1}{c|}{\multirow{2}{*}{\centering This article}}  \\ \cline{1-2}
AUC &    \citet{ling1998data,agarwal2005roc,menon2016bipartite}          &                               \\ \hline
\end{tabular}                           
\end{table*}

\subsection{Remark on other noise assumptions}
While this article focuses on the mutually contaminated noise assumption for learning from corrupted labels, it is worth noting that there are other noise assumptions used in binary classification. 
One of the most simple and well-studied noise assumptions is the random classification noise~\citep{oldccn,kearns1998efficient, long2010random, manwani2013noise, ghosh2015making, van2015learning}, which assumes that the observed noisy label can be flipped from the unobserved true label with the fixed probability $\sigma$ for any pattern $\x$, where $\sigma < 0.5$. 
This noise is known to be a special case of the mutually contaminated noise~\citep{menon2015}.
With the mutually contaminated noise assumption, it is possible to model the noise to be class-conditional, i.e., the probability of label flipping does not need to be a fixed constant but it can depend on the unobserved true label~$y$.
% It is also possible to use this assumption to formulate weakly supervised learning problems, as we will show in Section~\ref{sec:weakly}.
% In Sections~\ref{sec:symadvantage} and~\ref{sec:weakly}, we will show that this noise assumption can be used to guarantee the robustness of symmetric losses and to formulate weakly supervised learning problems.

There are other noise assumptions that could be considered more general than the mutually contaminated noise, e.g., the instance-dependent noise, where the probability of a label being mislabeled $\sigma_{\x, y}$ depends on both a pattern $\x$ and its unobserved true label~$y$~\citep{manwani2013noise,idn3,idn2, idn4, NeurIPS:Xia+etal:2020, idn1, CVPR:Cheng+etal:2022}. 
However, modeling this noise can be complex and it is typically required to have an additional restrictive assumption to provide an analysis of learning under this setting (e.g., see Table 1 in~\citet{idn1}).

While all noise assumptions mentioned above are used for modeling the behavior of noisy datasets in the real-world scenario, it is also worth pointing out that there are noise assumptions for other different purposes such as proving a better convergence rate of a learned classifier.
Massart noise assumption~\citep{massart2006risk} and Tsybakov noise assumption~\citep{mammen1999smooth,tsybakov2004optimal} are well-known examples, which are out of the scope of this article.

\section{A Symmetric Loss Approach to BER and AUC Optimization from Corrupted Labels}
\label{sec:symadvantage}
In this section, we begin by describing the related work in BER/AUC optimization from corrupted labels and then show that using a symmetric loss can be advantageous for BER and AUC optimization from corrupted labels. 

\subsection{Background}
In learning from corrupted labels, \citet{menon2015} proved that both BER and AUC are robust w.r.t. the zero-one loss. 
More precisely, without using a surrogate loss, the minimizer of the BER/AUC risk w.r.t. the clean distribution and that of the BER/AUC risk w.r.t. the corrupted distribution are identical.
In experiments, they used the squared loss as their choice of the surrogate loss and the comparison between the squared loss and other surrogate losses was not conducted.
Next, \citet{van2015average} generalized the theoretical result of \citet{menon2015} for BER minimization from corrupted labels from the zero-one loss to any symmetric loss.
Then, \citet{charoenphakdee2019symmetric} proved the relationship between the clean and corrupted BER/AUC risks for a general loss function, which elucidates that using a symmetric loss can be advantageous for both BER and AUC optimization from corrupted labels.
Furthermore, \citet{charoenphakdee2019symmetric} also conducted extensive experiments to verify that using symmetric losses can perform significantly better than using non-symmetric losses.  

\subsection{BER minimization from corrupted labels}
To verify the robustness of BER minimization from corrupted labels, we investigate the relationship between the clean risk with the corrupted risk for any surrogate loss~$\ell$.
First, let us define the following surrogate risk for BER from corrupted labels:
\begin{align*}
	R^\ell_{\mathrm{\widetilde{BER}}}(g) = \frac{1}{2} \big[ R_{\mathrm{\widetilde{P}}}^{\ell}(g) + R_{\mathrm{\widetilde{N}}}^{\ell}(g)  \big] \text{,}
\end{align*}
where 
\begin{align*}
R_{\mathrm{\widetilde{P}}}^{\ell}(g) &= \Eposc [\ell(g(\x))] =\pip \Epos[\ell(g(\x))]  + (1-\pip)\Eneg[\ell(g(\x))] \text{,} \\
R_{\mathrm{\widetilde{N}}}^{\ell}(g) &= \Enegc [\ell(-g(\x))] = \pin \Epos[\ell(-g(\x))]  + (1-\pin)\Eneg[\ell(-g(\x))]\text{,} 
\end{align*}
where $\Eposc$ and $\Enegc$ denote the expectations over $p_{\pip}$ and  $p_{\pin}$, respectively.
Since we have samples $X_{\mathrm{\widetilde{P}}}$ and $X_{\mathrm{\widetilde{N}}}$ following $p_{\pip}$ and $p_{\pin}$, respectively (see Section~\ref{sec:learncor}), the following empirical risk $\widehat{R}^\ell_{\mathrm{\widetilde{BER}}}$ can be minimized in practice:
\begin{align}
\label{eq:ber-emp}
    \widehat{R}^\ell_{\mathrm{\widetilde{BER}}}(g) = \frac{1}{2}\left[ \frac{1}{n_\mathrm{\widetilde{P}}}\sum_{\x \in X_{\mathrm{\widetilde{P}}}} \ell(g(\x)) + 
    \frac{1}{n_\mathrm{\widetilde{N}}}\sum_{\x \in X_{\mathrm{\widetilde{N}}}} \ell(-g(\x)) \right].
\end{align}
Note that $\widehat{R}^\ell_{\mathrm{\widetilde{BER}}}$ can be minimized without the knowledge of $\pip$ and $\pin$. 
Next, the following equation illustrates the relationship between the clean BER risk $R^\ell_{\mathrm{BER}}$ and the corrupted BER risk $R^\ell_{\mathrm{\widetilde{BER}}}$ w.r.t. to any loss function $\ell$~\citep{charoenphakdee2019symmetric}:
\begin{align}
\label{eq:ber-general}
R^\ell_{\mathrm{\widetilde{BER}}}(g) &= (\pip-\pin) {R^\ell_{\mathrm{BER}}(g)} + \underbrace{\frac{\pin \Epos[\gamma^\ell(g(\x))] + (1-\pip)\Eneg[\gamma^\ell(g(\x))]}{2}}_\textup{Excessive term} \text{,}
\end{align}
where $\gamma^\ell(z) = \ell(z) + \ell(-z)$.

From Eq.~(\ref{eq:ber-general}), we can see that $g$ which minimizes $R^\ell_{\mathrm{\widetilde{BER}}}$ should also perform reasonably well for  $R^\ell_{\mathrm{BER}}$ for any loss function.
However, together with $R^\ell_{\mathrm{BER}}$, a prediction function $g$ that minimizes $R^\ell_{\mathrm{\widetilde{BER}}}$ should also take the excessive terms into account, which are the terms that are unrelated to our goal. 
As a result, the minimizer of $R^\ell_{\mathrm{\widetilde{BER}}}$ may not be guaranteed to be the minimizer of $R^\ell_{\mathrm{BER}}$ because of the non-constant excessive term.

Next, let us consider a symmetric loss $\ell_\mathrm{sym}$ such that $\ell_\mathrm{sym}(z)+\ell_\mathrm{sym}(-z)=K$, where $K$ is a constant regardless of $z$. 
With a symmetric loss, we can rewrite Eq.~\eqref{eq:ber-general} as
\begin{align*}
    R^{\ell_\mathrm{sym}}_{\mathrm{\widetilde{BER}}}(g) &= (\pip-\pin) {R^{\ell_\mathrm{sym}}_{\mathrm{BER}}(g)} + \frac{\pin \Epos[K] + (1-\pip)\Eneg[K]}{2}\\
    &=   (\pip-\pin)  R^{{\ell_\mathrm{sym}}}_{\mathrm{BER}}(g) + K\left(\frac{1-\pip+\pin}{2}\right) \text{.}
\end{align*}

We can see that if a loss is symmetric, then the excessive term will be a constant and the minimizers of $R^\ell_{\mathrm{\widetilde{BER}}}$ and $R^\ell_{\mathrm{BER}}$ must be identical.
This suggests that $g$ can ignore the excessive term when using a symmetric loss.
As a result, BER minimization from corrupted labels can be done effectively without the knowledge of $\pip$ and $\pin$ by minimizing Eq.~\eqref{eq:ber-emp} using a symmetric loss. 

\subsection{AUC maximization from corrupted labels}\label{sec:aucmax}

Let us consider a corrupted AUC risk with a surrogate loss $\ell$ that treats $X_\mathrm{\widetilde{P}}$ as being positive and $X_{\mathrm{\widetilde{N}}}$ as being negative:
\begin{align*} 
R^\ell_{\mathrm{\widetilde{AUC}}}(g) = \mathbb{E}_\mathrm{\widetilde{P}}[\mathbb{E}_{\mathrm{\widetilde{N}}}[\ell( g(\x_\mathrm{\widetilde{P}})-g(\x_\mathrm{\widetilde{N}}))]]\text{,}
\end{align*}
which can be empirically approximated using training data as
\begin{align}
\label{eq:auc-emp}
        \widehat{R}^\ell_{\mathrm{\widetilde{AUC}}}(g) = \frac{1}{n_{\mathrm{\widetilde{P}}}\times n_{\mathrm{\widetilde{N}}} }\sum_{\x \in X_{\mathrm{\widetilde{P}}}} \sum_{\x^\prime \in X_{\mathrm{\widetilde{N}}}} \ell(g(\x)-g(\x^\prime))\text{.}
\end{align}
\citet{charoenphakdee2019symmetric} showed that the relationship between $R^\ell_{\mathrm{\widetilde{AUC}}}(g)$ and $R^\ell_{\mathrm{AUC}}(g)$ can be expressed as follows:
\begin{align*}
\begin{split}
R^\ell_{\mathrm{\widetilde{AUC}}}(g)  ={}& (\pip-\pin)R^\ell_{\mathrm{AUC}}(g)   + \underbrace{(1-\pip)\pin \Epos[\Eneg[\gamma^\ell(g(\x_\mathrm{P}),g(\x_\mathrm{N}))]]}_\textup{Excessive term} \\ &+ \underbrace{\frac{\pip\pin}{2} \EX_{\mathrm{P}}[\Epos[\gamma^\ell(g(\x'_{\mathrm{\mathrm{P}}}),g(\x_\mathrm{P}))]]}_\textup{Excessive term}  \\ &+ \underbrace{\frac{(1-\pip)(1-\pin)}{2}   \EX_{\mathrm{N}}[\Eneg[\gamma^\ell(g(\x'_{\mathrm{N}}),g(\x_\mathrm{N}))]]}_\textup{Excessive term}\text{,}
\end{split}
\end{align*}
where $\gamma^\ell(z,z') = \ell(z-z') + \ell(z'-z)$.
Next, with a symmetric loss $\ell_\mathrm{sym}$, we have
\begin{align*}
R^{\ell_\mathrm{sym}}_{\mathrm{\widetilde{AUC}}}(g)  &= (\pip-\pin)R^\ell_{\mathrm{AUC}}(g)   + (1-\pip)\pin \Epos[\Eneg[K]] \\ &\quad + \frac{\pip\pin}{2} \Epos[\Epos[K]]   + \frac{(1-\pip)(1-\pin)}{2}   \Eneg[\Eneg[K]]\text{,}
\\ 
&= (\pip-\pin)R^{\ell_\mathrm{sym}}_{\mathrm{AUC}}(g) +  K\left(\frac{1  -\pip + \pin}{2}\right) \text{.}
\end{align*}

Similarly to BER minimization from corrupted labels, the result suggests that the excessive terms become a constant when using a symmetric loss and it is guaranteed that the minimizers of $R^\ell_{\mathrm{\widetilde{AUC}}}(g)$ and $R^\ell_{\mathrm{AUC}}(g)$ are identical. 
On the other hand, if a loss is non-symmetric, then we may suffer from the excessive terms and the minimizers of both risks may differ.

We can see that in both BER and AUC optimization from corrupted labels, by using a symmetric loss, the knowledge of $\pip$ and $\pin$ are not required and we can treat the corrupted labels as if they were clean to learn in this setting. 
We refer the readers to~\citet{charoenphakdee2019symmetric} for more details on experimental results, where symmetric losses are shown to be preferable over non-symmetric losses. 

\section{BER and AUC Optimization for Weakly Supervised Learning Based on Learning from Corrupted Labels}
\label{sec:weakly}
In this section, we demonstrate that the formulation of BER and AUC optimization from corrupted labels can also be useful in weakly supervised learning by considering the problem of $(i)$ learning from positive of unlabeled data and $(ii)$ learning from two sets of unlabeled data with different class priors. 

\subsection{Learning from positive and unlabeled data}
Here, let us consider the problem of binary classification from positive and unlabeled data. 
We consider the case-control setting where the training data are given as follows~\citep{ward2sample,du2014,du2015convex,niu2016,kiryo2017,charoenphakdee2019positive,xu2019revisiting}:
\begin{align*}
 X_\mathrm{P}&:= \{\x^\mathrm{P}_i\}_{i=1}^{n_\mathrm{P}} \stackrel{\mathrm{i.i.d.}}{\sim} p(\x|y=+1) \text{,}\\
 X_{\mathrm{U}}&:=  \{\x^\mathrm{U}_j\}_{j=1}^{n_\mathrm{U}}\stackrel{\mathrm{i.i.d.}}{\sim} \pi_\mathrm{U} p(\x|y=+1)+(1-\pi_\mathrm{U}) p(\x|y=-1) \text{.}
\end{align*}
With $\pip=1$ and $\pin=\pi_\mathrm{U}$, we can relate the training data of learning from positive and unlabeled data to learning from corrupted labels.
In this setting,~\citet{sakai2018} showed that for AUC maximization, a convex surrogate loss can be applied but the class prior $\pi_\mathrm{U}$ needs to be estimated to construct an unbiased risk estimator. 
By using a symmetric loss, we can safely perform both BER and AUC optimization without estimating the class prior $\pi_\mathrm{U}$ with a theoretical guarantee.  

Concretely, with a symmetric loss $\sell$, BER minimization from positive and unlabeled data can be done effectively by minimizing the following empirical risk: 
\begin{align*}
    \widehat{R}^{\sell}_{\mathrm{BER}\text{-}\mathrm{PU}}(g) = \frac{1}{2}\left[ \frac{1}{n_\mathrm{P}}\sum_{\x \in X_{\mathrm{P}}} \sell(g(\x)) + 
    \frac{1}{n_\mathrm{U}}\sum_{\x^\prime \in X_{\mathrm{U}}} \sell(-g(\x^\prime)) \right],
\end{align*}
and AUC maximization can be done effectively by minimizing the following empirical risk:
\begin{align*}
        \widehat{R}^{\sell}_{\mathrm{AUC}\text{-}\mathrm{PU}}(g) = \frac{1}{n_{\mathrm{P}}\times n_{\mathrm{U}} }\sum_{\x \in X_{\mathrm{P}}} \sum_{x^\prime \in X_{\mathrm{U}}} \sell(g(\x)-g(x^\prime))\text{.}
\end{align*}

\subsection{Learning from two sets of unlabeled data with different class priors}
Here, let us consider the problem of binary classification from two set of unlabeled data, where the training data are  given as follows~\citep{lu2018minimal,lu2020mitigating}: 
\begin{align*}
 X_\mathrm{U}&:= \{\x^\mathrm{U}_i\}_{i=1}^{n_\mathrm{U}} \stackrel{\mathrm{i.i.d.}}{\sim} \pi_\mathrm{U} p(\x|y=+1)+(1-\pi_\mathrm{U}) p(\x|y=-1) \text{,}\\
 X_{\mathrm{U'}}&:=  \{\x^\mathrm{U'}_j\}_{j=1}^{n_\mathrm{U'}}\stackrel{\mathrm{i.i.d.}}{\sim} \pi_\mathrm{U'} p(\x|y=+1)+(1-\pi_\mathrm{U'}) p(\x|y=-1) \text{,}
\end{align*}
where $\pi_\mathrm{U} > \pi_\mathrm{U'}$.

We can relate given training data of this problem to learning from corrupted labels by having $\pip=\pi_\mathrm{U}$ and $\pin=\pi_\mathrm{U'}$.
Therefore, BER and AUC optimization from two sets of unlabeled data with different class priors can also be carried out effectively with a symmetric loss without knowing class priors $\pi_\mathrm{U}$ and $\pi_\mathrm{U'}$.
It is interesting to see that although the data collection procedure of learning from corrupted labels and learning from two sets of unlabeled data are very different in practice, the assumptions of the data generating process can be highly related.

Concretely, with a symmetric loss $\sell$, BER minimization from two sets of unlabeled data can be done effectively by minimizing the following empirical risk: 
\begin{align*}
    \widehat{R}^{\sell}_{\mathrm{BER}\text{-}\mathrm{UU}}(g) = \frac{1}{2}\left[ \frac{1}{n_\mathrm{U}}\sum_{\x \in X_{\mathrm{U}}} \sell(g(\x)) + 
    \frac{1}{n_\mathrm{U'}}\sum_{\x' \in X_{\mathrm{U'}}} \sell(-g(\x')) \right],
\end{align*}
and AUC maximization can be done effectively by minimizing the following empirical risk:
\begin{align*}
        \widehat{R}^{\sell}_{\mathrm{AUC}\text{-}\mathrm{UU}}(g) = \frac{1}{n_{\mathrm{U}}\times n_{\mathrm{U'}} }\sum_{\x \in X_{\mathrm{U}}} \sum_{x^\prime \in X_{\mathrm{U'}}} \sell(g(\x)-g(x^\prime))\text{.}
\end{align*}

\section{A Symmetric Loss Approach to Learning Only from Relevant Keywords and Unlabeled Documents}\label{sec:application}
In this section, we demonstrate how to apply the robustness result of symmetric losses in learning from corrupted labels to tackle a natural language processing task, namely learning only from relevant keywords and unlabeled documents~\citep{charoenphakdee2019learning}. 

\subsection{Background}
To reduce the labeling costs, weakly supervised text classification has been studied extensively in various settings, e.g., positive and unlabeled text classification~\citep{li2003learning,li2005learning}, zero-shot text classification~\citep{zhang2019integrating}, cross-lingual text classification~\citep{dong2019robust}, and dataless classification~\citep{chang2008importance,song2014dataless,chen2015dataless,jin2017combining,jin2020learning,li2018pseudo}.
Our target problem can be categorized as a variant of dataless classification, where we are given a set of $n_\mathrm{W}$ relevant keywords:
\begin{align*}
W:=  \{w_j\}_{j=1}^{n_\mathrm{W}},
\end{align*}
which are keywords that provide characteristics of positive documents. 
Also, unlabeled documents drawn from the following distribution are provided:
\begin{align*}
    X_{\mathrm{U}}:=  \{\x^\mathrm{U}_i\}_{i=1}^{n_\mathrm{U}}\stackrel{\mathrm{i.i.d.}}{\sim} p_{\pi_\mathrm{U}}(\x),
\end{align*}
where, for $\pi_\mathrm{U} \in (0,1)$,
\begin{align*}
    p_{\pi_\mathrm{U}}(\x) = \pi_\mathrm{U} p(\x|y=+1) +(1-\pi_\mathrm{U})\, p(\x|y=-1).
\end{align*}
Note that unlike ordinary dataless classification, where we need keywords for every class~\citep{chang2008importance,song2014dataless,chen2015dataless,li2018pseudo}, in this problem, only keywords for the positive class are provided.
Therefore, this problem setting could be more practical in a situation where negative documents are too diverse to collect representative keywords for the negative class~\citep{hsieh2019classification}.

It is worth noting that our problem is also called \emph{lightly-supervised learning}~\citep{jin2017combining}, where the supervision comes from the relevant keywords.
To solve this problem, \citet{jin2017combining} proposed to use a method based on ensemble learning.
The bottleneck of the method proposed by~\citet{jin2017combining} is lack of flexibility of model choices and optimization algorithms.
This makes it difficult to bring many useful models and techniques from a more well-studied problem such as supervised text classification to help solve this problem.
Moreover, the theoretical understanding of this problem was limited.
To alleviate these limitations, \citet{charoenphakdee2019learning} proposed a theoretically justifiable framework that allows practitioners to choose their preferred models to maximize the performance, e.g., convolutional neural networks~\citep{zhang2015character} or recurrent neural networks~\citep{lai2015recurrent}.
Moreover, this framework does not limit the choice of optimization methods. 
One may use any optimization algorithm for their model, e.g., Adam~\citep{adam}.

In learning only from relevant keywords and unlabeled documents, the choice of evaluation metrics depends on the desired behavior of the prediction function $g$ we want to learn.
For example, AUC is appropriate if the goal is simply to learn a bipartite ranking function to rank a positive document over a negative document.
On the other hand, if the goal is document classification, one may use CER or the $\mathrm{F}_{1}$-measure, i.e., the harmonic mean of precision and recall, which has been widely used in text classification~\citep{li2005learning,jin2017combining,jin2020learning,lertvittayakumjorn2019human,lertvittayakumjorn2020find,mekala2020contextualized,he2020towards}. 

\subsection{A flexible framework for learning only from relevant keywords and unlabeled documents}
\begin{figure}[t]
\vspace{0.1in}
\begin{center}
 \centerline{\includegraphics[width = \textwidth]{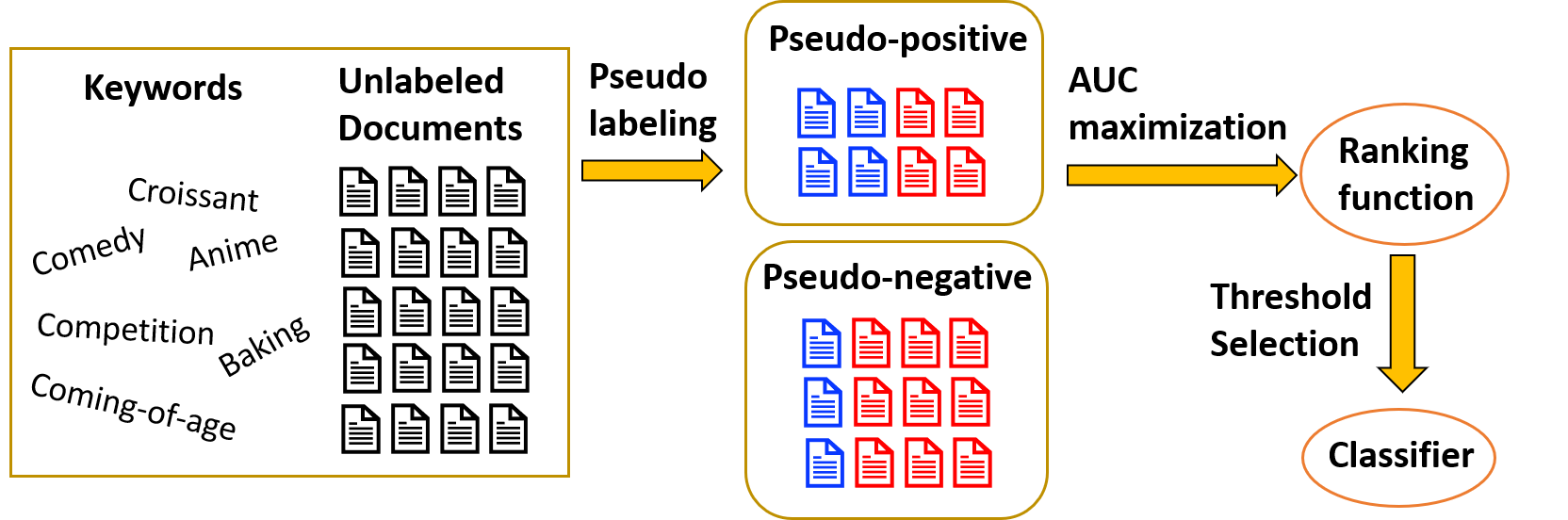}}
\caption{An overview of the framework for learning only from relevant keywords and unlabeled documents~\citep{charoenphakdee2019learning}. 
Blue documents indicate positive documents and red documents denote negative documents in the two sets of documents divided by a pseudo-labeling algorithm. 
Note that clean labels are not observed by the framework.}
\label{fig:ete-network}
\end{center}
\vspace{-0.35in}
\end{figure}

Here, we discuss a flexible framework for learning only from relevant keywords and unlabeled documents proposed by~\citet{charoenphakdee2019learning}.
Figure~\ref{fig:ete-network} illustrates an overview of the framework.
First, pseudo-labeling is carried out to split unlabeled documents into two sets.
Next, by using pseudo-labeled documents, AUC maximization is conducted, where the choice of the surrogate loss is a symmetric loss.
Finally, after obtaining a bipartite ranking function by AUC maximization, a threshold selection procedure is performed to convert the ranking function to a binary classifier.

\subsubsection{Pseudo-labeling: from learning with keywords and unlabeled documents to learning from corrupted labels}\label{sec:pseudo-label}
The first step is to utilize relevant keywords to perform pseudo-labeling on unlabeled documents.
Concretely, given relevant keywords $W$ and unlabeled documents~$X_\mathrm{U}$, the pseudo-labeling algorithm $A(W,X_\mathrm{U})$ splits $X_\mathrm{U}$ into $X_\mathrm{\widetilde{P}}$ and $X_\mathrm{\widetilde{N}}$.
The key idea of this step is to use pseudo-labeling to bridge learning only from relevant keywords and unlabeled documents to learning from corrupted labels. 
More precisely, we assume that the pseudo-labeling algorithm $A(W,X_\mathrm{U})$ returns the following data:
\begin{align*}
 X_\mathrm{\widetilde{P}}&:= \{\x^\mathrm{\widetilde{P}}_i\}_{i=1}^{n_\mathrm{\widetilde{P}}} \stackrel{\mathrm{i.i.d.}}{\sim} p_{\pip}(\x) \text{,}\\
 X_{\mathrm{\widetilde{N}}}&=  \{\x^\mathrm{\widetilde{N}}_j\}_{j=1}^{n_\mathrm{\widetilde{N}}}\stackrel{\mathrm{i.i.d.}}{\sim} p_{\pin}(\x) \text{,}
\end{align*}
where the assumption of the data generating process is identical to that of the setting of learning from corrupted labels (see Section~\ref{sec:learncor}). 

It is important to note that a pseudo-labeling algorithm we employ here is not expected to perfectly split documents into clean positive and negative documents. 
For the choice of the pseudo-labeling algorithm, ~\citet{charoenphakdee2019learning} simply used a cosine similarity score between keywords and documents and compared the score with a pre-defined threshold to split unlabeled documents into two sets. 
To further improve the pseudo-labeling accuracy, one may utilize domain-specific knowledge or a keyword mining method to collect more relevant keywords.
Examples of such keyword mining methods are Textrank~\citep{mihalcea2004textrank} and Topicrank~\citep{bougouin2013topicrank}.
Moreover, one may also incorporate an unsupervised learning method~\citep{ko2000automatic, ko2009text} or apply a pre-trained model such as BERT~\citep{devlin2018bert}.

\subsubsection{AUC maximization from pseudo-labeled documents: obtaining a reliable ranking function from unreliable data}
After pseudo-labeling, we can conduct AUC maximization using a symmetric loss to learn a good ranking function $g$ from pseudo-labeled documents.

Recall that, with any symmetric loss, the AUC risk minimizers of the corrupted risk and clean risk are identical, which is suggested from the following equation:
\begin{align}
\label{eq:same-minimizer}
R^{\ell_\mathrm{sym}}_{\mathrm{\widetilde{AUC}}}(g) = (\pip-\pin)R^{\ell_\mathrm{sym}}_{\mathrm{AUC}}(g) +  K\left(\frac{1  -\pip + \pin}{2}\right) \text{.}
\end{align}
Eq.~\eqref{eq:same-minimizer} indicates that as long as the pseudo-labeling algorithm succeeds in splitting the documents into two sets such that $\pip > \pin$, we can always guarantee that~$g$ can be effectively learned from pseudo-labeled documents. 
More precisely, the minimizers of the risk w.r.t. the pseudo-labeled document distribution and the clean document distribution are identical.
However, since any pseudo-labeling algorithm that gives $\pip > \pin$ can guarantee to learn a good ranking function, one important question is: how does the quality of the pseudo-labeling method impact the performance of the trained prediction function $g$ in this framework?
Intuitively, a good pseudo labeling should give a high proportion of positive documents in the pseudo-positive set and a high proportion of negative documents in the pseudo-negative set.
Mathematically, a good algorithm should return two sets of documents with a large $\pip$ and a small $\pin$, that is, $\pip-\pin$ is large.

To elucidate the usefulness of a good pseudo-labeling algorithm, it is insightful to analyze an estimation error bound of AUC maximization from corrupted labels. 
Let~$\hat{g} \in \mathcal{G}$ be a minimizer of the empirical corrupted AUC risk $\widehat{R}^{\sell}_{\mathrm{\widetilde{AUC}}}$ in the hypothesis class $\mathcal{G}$ and $g^* \in \mathcal{G}$ be a minimizer of the clean AUC risk $R^{\sell}_{\mathrm{AUC}}$.
Then, the following theorem can be obtained.
\begin{theorem}[Estimation error bound~\citep{charoenphakdee2019learning}]\label{est-bound}  
Let $\mathcal{Q}_\mathcal{G}^{\sell}$ be a class of functions mapping $\X^2$ to $[0,K]$ such that $\mathcal{Q}_\mathcal{G}^{\sell}= \{Q: (\x, \x') \to \sell(g(\x)-g(\x')), g \in \mathcal{G}\}$.
$\widetilde{\mathfrak{R}}_{n_\mathrm{\widetilde{P}}, n_\mathrm{\widetilde{N}}}(\mathcal{Q}_\mathcal{G}^{\sell})$ denotes the bipartite Rademacher complexities of function class $\mathcal{Q}_\mathcal{G}^{\sell}$ (see ~\citet{usunier2005data} for more details).
For all $Q \in \mathcal{Q}_\mathcal{G}^{\sell}$ and $\delta \in$ (0,1), with probability at least $1-\delta$, we have
\begin{align*}
   R^{\sell}_{\mathrm{AUC}}(\hat{g})-R^{\sell}_{\mathrm{AUC}}(g^*) \leq   \frac{1}{\pip-\pin} \bigg[ K \sqrt{\frac{2(n_\mathrm{\widetilde{P}}+n_\mathrm{\widetilde{N}})\log\frac{1}{\delta}}{n_\mathrm{\widetilde{P}}n_\mathrm{\widetilde{N}}}}  + 4\widetilde{\mathfrak{R}}_{n_\mathrm{\widetilde{P}}, n_\mathrm{\widetilde{N}}}(\mathcal{Q}_\mathcal{G}^{\sell})\bigg],
\end{align*}
where the probability is over repeated sampling of $X_\mathrm{\widetilde{P}}$ and $X_\mathrm{\widetilde{N}}$.
\end{theorem}

This theorem explains how the degree of corruption $\pip-\pin$ affects the tightness of the bound and therefore the speed of convergence.
When $\pip-\pin$ is small, i.e., $\pip$ and $\pin$ are similar, the bound becomes loose. 
This illustrates the difficulty of AUC maximization when the pseudo-labeling algorithm performs poorly and we may need a lot of data. 
On the other hand, a good pseudo-labeling that gives a large $\pip-\pin$ can give a smaller constant $\frac{1}{\pip-\pin}$, which can lead to a tighter bound.
Nevertheless, it is noteworthy that as long as $\pip > \pin$, with all parametric models having a bounded norm such as neural networks with weight decay or kernel models, this learning framework is \textit{statistically consistent}, i.e., the estimation error converges to zero as $n_\mathrm{\widetilde{P}},n_\mathrm{\widetilde{N}} \to \infty$. 
\par

\subsubsection{Threshold selection: from a ranking function to a binary classifier}
After obtaining a good ranking function $g$, an important question is how to convert the ranking function to a binary classifier.
Here, we discuss how to decide the classification threshold for optimizing evaluation metrics such as the $\mathrm{F}_{1}$-measure.

It is known that many evaluation metrics can be optimized if a suitable threshold and $p(y=+1|\x)$ are known~\citep{yan2018binary}. 
For example, $\mathrm{sign}[p(y=+1|\x)-\frac{1}{2}]$ is the Bayes-optimal solution for the classification accuracy, where $\frac{1}{2}$ is the threshold.
Moreover, it has been proven that the Bayes-optimal solution of AUC maximization with an AUC-consistent surrogate loss is any function that has a strictly monotonic relationship with $p(y=+1|\x)$~\citep{clemenccon2009tree,gao2015consistency,menon2016bipartite}.
Therefore, finding an appropriate threshold to convert a bipartite ranking function to a binary classifier can give a reasonable classifier~\citep{narasimhan2013relationship}.

In learning from relevant keywords and unlabeled documents, no information about a proper threshold can be obtained from the training data since all given data are unlabeled.
For this reason, we may not be able to draw an optimal threshold for optimizing the accuracy and $\mathrm{F}_{1}$-measure without additional assumptions.
On the other hand, as shown in Section~\ref{sec:aucmax}, a ranking function can be learned reliably with a theoretical guarantee based on AUC maximization.
This is the main reason why ~\citet{charoenphakdee2019learning} proposed to first learn a reliable ranking function instead of learning a binary classifier in this problem.

Suppose the class prior $p(y=+1)$ in unlabeled documents is known, a reasonable threshold~$\beta \in \R$ can be given based on the following equation:
\begin{align}\label{eq:thr}
    p(y=+1) = \int \mathrm{sign}(g(\x)-\beta) p_{\pi_\mathrm{U}}(\x) \mathrm{d}\x.
\end{align}

Intuitively, the threshold $\beta$ allows $g$ to classify the top proportion of unlabeled documents w.r.t. $p(y=+1|\x)$ to be positive, and negative otherwise.
With unlabeled documents and the known proportion of positive documents, one can decide $\beta$ that satisfies the empirical version of Eq.~\eqref{eq:thr}. 
Concretely, given unlabeled documents $\X_{\mathrm{val}}$ for validation, the threshold can be decided by finding $\beta$ such that
\begin{align*}
    p(y=+1) \approx \frac{1}{|\X_{\mathrm{val}}|}\sum_{\x \in \X_{\mathrm{val}}} \mathrm{sign}(g(\x)-\beta).
\end{align*}

This threshold is known as a precision-recall breakeven point, where it is the point that the precision is equal to recall (see~\citet{kato2018learning} for its proof). 
Therefore, this choice is arguable to be a reasonable threshold for the $\mathrm{F}_{1}$-measure since this evaluation metric is the harmonic mean of precision and recall. 
In practice, it may not be possible to know the proportion $p(y=+1)$, yet we still want a classifier.
Without knowing the proportion of positive documents, it is likely that we learn a wrong threshold, which leads to low performance.
For example, as shown in Table~\ref{tab:failure}, the performance degrades dramatically with a wrong choice of the threshold.
More details on a heuristic for threshold selection and the performance w.r.t. different thresholds can be found in~\citet{charoenphakdee2019learning}.

\begin{table}[t]
\centering
\caption{Mean value and standard error of 20 trials with different thresholds on the Subjectivity dataset ($\mathsf{Subj}$)~\citep{Pang+Lee:04a} and the 20 Newsgroups dataset ($\mathsf{20NG}$)~\citep{Lang95} with the \emph{baseball} and \emph{hockey} groups as positive. 
ACC denotes the classification accuracy and $\mathrm{F}_{1}$ denotes the $\mathrm{F}_{1}$-measure. 
``Sigmoid'' is the framework using the sigmoid loss~\citep{charoenphakdee2019learning}, which employs a recurrent convolutional neural networks model (RCNN)~\citep{lai2015recurrent} with two layer long short-term memory (LSTM)~\citep{hochreiter1997long}. 
The ``heuristic threshold'' is the ratio between pseudo-positive documents over unlabeled documents 
and the ``default threshold'' is a baseline choice (see~\citet{charoenphakdee2019learning} for details).
It can be seen that given the same ranking function, the classification performance can drastically change depending on the choice of the threshold.} \label{tab:failure}
\resizebox{\columnwidth}{!}{\begin{tabular} { |c|c|c|c|c|c|c|c|c|}
\hline
\multirow{2}{*}{Method}& \multirow{2}{*}{Dataset}&  \multirow{2}{*}{AUC}& \multicolumn{2}{c|}{Default threshold} & \multicolumn{2}{c|}{Heuristic threshold} & \multicolumn{2}{c|}{True threshold}\\ 
& & & $\mathrm{F}_{1}$ & ACC & $\mathrm{F}_{1}$ & ACC & $\mathrm{F}_{1}$ & ACC \\ 
\hline
\multirow{2}{*}{Sigmoid}& $\mathsf{Subj}$ & 88.1 (0.35) & 69.0 (1.75) & 71.4 (1.33) & 80.1 (0.35) & 80.2 (0.35) & 80.1 (0.38) & 80.1 (0.38) \\
& $\mathsf{20NG}$ & 96.4 (0.12) & 59.0 (0.92) & 68.4 (1.19) & 83.8 (0.24) & 95.1 (0.06) & 90.8 (0.20) & 96.5 (0.08) \\
\hline
\multirow{2}{*}{Maxent}& $\mathsf{Subj}$ & 84.6 (0.20) & 63.4 (0.31) & 66.9 (0.23) & 76.8 (0.21) & 76.8 (0.21) & 76.3 (0.24) & 76.3 (0.24) \\
& $\mathsf{20NG}$ & 57.6 (0.32) & 47.4 (0.05) & 89.2 (0.02) & 51.8 (0.22) & 85.1 (0.09) & 52.4 (0.25) & 81.6 (0.11) \\
\hline
\multirow{2}{*}{RandomForest}& $\mathsf{Subj}$ & 82.4 (0.27) & 33.3 (0.00) & 50.00 (0.00) & 54.8 (0.13) & 60.9 (0.09) & 75.1 (0.27) & 75.1 (0.27) \\
& $\mathsf{20NG}$ & 96.8 (0.16) & 47.2 (0.00) & 89.4 (0.00) & 83.2 (0.32) & 95.0 (0.08) & 89.6 (0.28) & 96.1 (0.10) \\
\hline 
\end{tabular}}
\end{table}

It is also worth noting that if we cannot guarantee that $\pi_\mathrm{U}\approx p(y=+1|\x)$, i.e., the proportion of positive documents in the unlabeled training documents is similar to that of the test data, then using a metric such as the $\mathrm{F}_{1}$-measure or classification accuracy is highly biased to the training distribution~\citep{scott2012calibrated,narasimhan2014statistical}.
This problem is known as the class prior shift phenomenon and it might occur in real-world applications~\citep{saerens2002adjusting, datashift,tasche2017fisher}.
For example, by collecting unlabeled documents from the internet, the proportion of positive documents can be different from that of the test distribution where we want to deploy the trained classifier.
Note that the class prior shift phenomenon can dramatically degrade the performance of a classifier~\citep{saerens2002adjusting, datashift, charoenphakdee2019positive}.
Therefore, if it is impossible to estimate the class prior $p(y=+1|\x)$ or the test environment is susceptible to the class prior shift phenomenon, we suggest to use other evaluation metrics such as BER or the AUC score.

\section{Conclusions}
\label{sec:discussion} 
In this article, we have reviewed recent advances in reliable machine learning from a symmetric loss perspective. 
We showed in Section~\ref{sec:symadvantage} that if a symmetric loss is used for BER and AUC optimization from corrupted labels, the corrupted and clean risks have the same minimizer regardless of the model.
In Section~\ref{sec:weakly}, we further showed that the robustness result in learning from corrupted labels can also be applied in BER and AUC optimization in learning from positive and unlabeled data, and learning from two sets of unlabeled data with different class priors.
Moreover, we demonstrated in Section~\ref{sec:application} that the theoretical advantage of symmetric losses is also practically valuable in learning only from relevant keywords and unlabeled documents. 
In this section, we conclude this article by discussing two future directions for the symmetric loss perspective of reliable machine learning.

The first direction is exploring more applications of symmetric losses for reliable machine learning.
Here, we provide two examples of this direction.
First, it has been recently shown that using a symmetric loss can also be beneficial in imitation learning from noisy demonstrations, where the goal is to teach an agent to imitate expert demonstrations although training data contain both expert and non-expert demonstrations. 
~\citet{tangkaratt2020robust} showed that imitation learning with a symmetric loss can enable an agent to successfully imitate expert demonstrations without assuming a strong noise assumption such as it is Gaussian~\citep{tangkaratt2020variational} or requiring additional confidence scores for given demonstrations~\citep{wu2019imitation,brown2019extrapolating,brown2020safe}. 
Another example is to use a symmetric loss in classification with rejection, where a classifier is allowed to refrain from making a prediction if the prediction is uncertain~\citep{chow1970,yuan2010,cortes2016learning,ni2019calibration,mozannar2020consistent}.
Although well-known symmetric losses have favorable properties such as classification-calibration and AUC-consistency, it is important to note that learning with a symmetric loss cannot guarantee to give a classifier with reliable prediction confidence~\citep{charoenphakdee2019symmetric}.
Recently, ~\citet{charoenphakdee2020classification} proposed an approach based on cost-sensitive classification, which enables any classification-calibrated loss including symmetric losses to be applied for solving classification with rejection.
In the experiments, the sigmoid loss was shown to be highly preferable in classification with rejection from noisy labels and classification with rejection from positive and unlabeled data.
These examples emphasize the potential of symmetric losses for reliable machine learning in addition to what we have introduced in this article.%Sections~\ref{sec:symadvantage} and \ref{sec:application}.
Therefore, it could be useful to explore the use of symmetric losses for a wider range of problems, e.g., domain adaptation~\citep{sugiyama2012machine,ben2012minimizing,kuroki2019unsupervised,lee2019domain,redko2020survey}, open-set classification~\citep{saito2018open,ruff2020unifying,fang2020open,geng2020recent}, and learning from aggregate observations~\citep{maron1997framework,hsu2019multi,cui2020classification,zhang2020learning}.

Although symmetric losses are useful in learning from corrupted labels, using them can sometimes lead to undesirable performance because training with a symmetric loss can be computationally hard.~\citep{ghosh2017robust,wang2019symmetric,ma2020normalized}.
Thus, it is interesting to explore non-symmetric losses that can benefit from the symmetric condition.
This is the second future direction we discuss in this section.
Here, we provide two examples to demonstrate the potential of this research direction.
The first example is motivated by the fact that a nonnegative symmetric loss must be non-convex~\citep{du2014}.
To explore a robust convex loss,~\citet{charoenphakdee2019symmetric} proposed the barrier hinge loss, which is a convex loss that satisfies a symmetric condition on a subset of the domain, but not everywhere.
The barrier hinge loss was shown to be highly robust in BER and AUC optimization although it is not symmetric. 
This suggests that one can design a non-symmetric loss that benefits from the symmetric condition.
Another example is an approach to combine a symmetric loss with a non-symmetric loss.
Recently,~\citet{wang2019symmetric} proposed the reverse cross-entropy loss, which is a symmetric loss.
Then, they proposed to combine the reverse cross-entropy loss with the ordinary cross-entropy loss by linear combination. 
Their experiments showed that the classification performance of the combined loss can be better than using only the reverse cross-entropy loss or other symmetric losses such as the mean absolute error loss~\citep{ghosh2017robust}.
Based on these examples, we can see that it could be beneficial to design a loss function that enjoys the advantages of both symmetric and non-symmetric losses.

\begin{acknowledgement}
We would like to thank our collaborators: Dittaya Wanvarie, Yiping Jin, Zhenghang Cui, Yivan Zhang, and Voot Tangkaratt.
NC was supported by MEXT scholarship and Google PhD Fellowship program.
MS was supported by JST CREST Grant Number JPMJCR18A2.
\end{acknowledgement}


\bibliography{ref}    % name your BibTeX data base
\bibliographystyle{spbasic}      % basic style, author-year citations

%---%
%\bibliographystyle{plainnat}

%\begin{thebibliography}{}
% etc
%\end{thebibliography}
\medskip
\end{document}